\documentclass{article}
\usepackage[utf8]{inputenc}
\usepackage{natbib}
\usepackage{titlesec}
\setcitestyle{authoryear,open={(},close={)}}
\usepackage{amssymb}
\usepackage{titlesec}
\usepackage{amsmath}
\usepackage{authblk}
\usepackage{array}
\usepackage{multirow}
\usepackage{graphicx}
\usepackage{enumitem}

\titleformat{\paragraph}
{\normalfont\normalsize\bfseries}{\theparagraph}{1em}{}
\titlespacing*{\paragraph}
{0pt}{3.25ex plus 1ex minus .2ex}{1.5ex plus .2ex}

\title{Deflating Deflationism: A Critical Perspective on Debunking Arguments Against LLM Mentality}

\author[1,*]{\normalsize Alex Grzankowski}
\author[2,*]{Geoff Keeling} 
\author[3,*]{Henry Shevlin}
\author[2,*]{Winnie Street}

\affil[1]{\footnotesize Institute of Philosophy, University of London}
\affil[2]{Google, Paradigms of Intelligence Team}
\affil[3]{Leverhulme Centre for the Future of Intelligence, University of Cambridge}
\affil[*]{Equal contributions, authors listed alphabetically}

\date{\footnotesize June 2025}

\begin{document}

\maketitle

\begin{abstract}
Many people feel compelled to interpret, describe, and respond to Large Language Models (LLMs) as if they possess inner mental lives similar to our own. Responses to this phenomenon have varied. \textit{Inflationists} hold that at least some folk psychological ascriptions to LLMs are warranted. \textit{Deflationists} argue that all such attributions of mentality to LLMs are misplaced, often cautioning against the risk that anthropomorphic projection may lead to misplaced trust or potentially even confusion about the moral status of LLMs. We advance this debate by assessing two common deflationary arguments against LLM mentality. What we term the \textit{robustness strategy} aims to undercut one justification for believing that LLMs are minded entities by showing that putatively  cognitive and humanlike behaviours are not robust, failing to generalise appropriately. What we term the \textit{etiological strategy} undercuts attributions of mentality by challenging naive causal explanations of LLM behaviours, offering alternative causal accounts that weaken the case for mental state attributions. While both strategies offer powerful challenges to full-blown inflationism, we find that neither strategy provides a knock-down case against ascriptions of mentality to LLMs \textit{simpliciter}. With this in mind, we explore a modest form of inflationism that permits ascriptions of mentality to LLMs under certain conditions. Specifically, we argue that folk practice provides a defeasible basis for attributing mental states and capacities to LLMs provided those mental states and capacities can be understood in metaphysically undemanding terms (e.g. knowledge, beliefs and desires), while greater caution is required when attributing metaphysically demanding mental phenomena such as phenomenal consciousness.
\end{abstract}

\section{Introduction}

It is hard to shake the thought that Large Language Models (LLMs) have minds. When interacting with LLM-based dialogue agents like ChatGPT, Claude and Gemini, many users ask these systems for advice, joke with them, and confide in them. Some even fall in love. Accordingly, users often find themselves speaking to LLMs as if these systems have thoughts, desires, or points of view. Is the attribution of mentality to LLMs a mistake---an anthropomorphic glitch in our social cognition? Or could it be reasonable to treat LLMs as minded?\footnote{\citet{colombatto2024folk}, \citet{ibrahim2025multi} and \citet{scott2023you} provide empirical assessments of user perceptions of mentality in LLMs. For ethical discussions of anthropomorphism in relation to LLMs see \citet{weidinger2021ethical}, \citet{gabriel2024ethics}, \citet{shevlin2021non}, \citet{manzini2024code}, \citet{manzini2024should}, \citet{akbulut2024all} and \citet{reinecke2025double}. For further discussion see \citet{guingrich2023chatbots}.}

The question of whether and when it is appropriate to ascribe mentality to LLMs has become a central flashpoint in the societal conversation on AI. While some researchers are engaged in exploratory work evaluating models for mental phenomena such as preferences, beliefs, and pleasure and pain states \citep{Anthropic2025Claude4, scherrer2023evaluating, keeling2024can}, others remain deeply sceptical \citep{Li_Etchemendy_2024, bender2021dangers}. We here distinguish two sets of views about LLM mentality which we term  \textit{inflationism} and \textit{deflationism}. Inflationism holds  that at least some attributions of mentality to LLMs are accurate or appropriate, while deflationism denies this.\footnote{\citet{palminteri2025navigating} also use the term `inflationism.' According to them, `[t]he inflationary stance consists of easily attributing higher-order cognition to LLMs, often extrapolating from their linguistic fluency to claims about Artificial General Intelligence (AGI) or even artificial consciousness' (see also \citeauthor{maclure2020new}, \citeyear{maclure2020new}). Our usage of the term `inflationism' is broader than theirs in that inflationism as we understand it encompasses any ascriptions of mentality to LLMs---not merely ascriptions of higher-order cognition and consciousness.} In framing the debate in these terms, we aim to draw a broad heuristic distinction, rather than a strict conceptual demarcation: in practice, commentators are often not inflationist or deflationist \textit{tout court}, but instead focus their arguments on some particular mental capacity or class of mental states.\footnote{\citet{CappelenManuscript-CAPGWH} are an exception. Their book manuscript \textit{Going the Whole Hog: A Philosophical Defense of AI Cognition} provides a broad defence of inflationism, although the form of inflationism that they defend is neutral on the issue of consciousness.} For example, \citet{bender2020climbing} defend deflationism with respect to semantic understanding \citep[c.f.][]{bender2021dangers}, and \citet{sogaard2023grounding} defends the corresponding inflationist view.

We aim to step back from these debates and take a broader methodological perspective, examining two general deflationist strategies for debunking mental state attributions to LLMs: the \textit{robustness strategy}, which challenges mental states attributions on a broadly functional basis, emphasising failures of generalisation; and the \textit{etiological strategy}, which appeals to the causal history of LLMs to undermine the case for interpreting their behaviour in mentalistic terms.

We argue that both strategies fall short of decisively undermining inflationism, with a recurring problem being that they beg the question against inflationist alternatives. Once these debunking arguments are defused, a modest form of inflationism emerges as not only defensible but perhaps even the natural starting point in interpreting the behaviour of LLMs. In defending modest inflationism, our aim is not to claim that LLMs have rich mental lives, or that their internal workings closely resemble those of human minds. Rather, we suggest that some amount of mentalising is sometimes justified. In particular, where the mental states or capabilities at issue can be understood in metaphysically undemanding terms (as has been argued for the case of beliefs and desires), in contrast to more metaphysically demanding states such as phenomenally conscious experiences. Accordingly, the modest inflationism that we endorse registers that ascribing mentality to LLMs need not entail ascribing mental states and capacities which are \textit{equivalent} to those of humans---in much the same way that ascriptions of beliefs, desires and intentions to non-human animals need not presuppose that their realisations of those mental states are exactly equivalent to ours. 

We note also that in what follows, our primary focus will be specifically on LLM-based dialogue agents such as ChatGPT, Claude, and Gemini \citep{shanahan2024talking}. While much of what we say will generalise to debunking arguments as applied to AI systems in general, we believe it is useful to have a specific target in mind, not least because the sheer variety of actual and possible AI systems (including neuromorphic architectures or even systems with biological components) introduces nuance and complexity beyond the scope of this paper.

The paper proceeds as follows. In Section 2, we lay out a basic plausibility case for mentalising LLMs. Sections 3 and 4 critically examine the robustness and etiological strategies as deflationary attempts to debunk this plausibility. In Section 5, we revisit the inflationism/deflationism debate in light of these arguments, and sketch a moderate, tractable form of inflationism that countenances attributions of mentality to LLMs provided the relevant attributions are in a relevant way metaphysically undemanding. The upshot is that the case for mentalising AI may be stronger than often assumed.

\section{Ascribe Mentality, Absent Defeaters}

LLMs bear many of the hallmarks of minded entities \citep{akbulut2024all, manzini2024code, gabriel2024ethics, CappelenManuscript-CAPGWH}. For example, LLMs can (or at least appear to) ask and answer questions, reason through complex problems, give advice, and make jokes. Humans naturally interpret behaviours like these as evidence for mentality---be it mental states such as beliefs, desires and intentions, or mental capacities such as phenomenal consciousness and semantic understanding. It is therefore unsurprising that many people who interact with LLMs ascribe LLMs some degree of mentality \citep{scott2023you, colombatto2024folk}.\footnote{Philosophers are increasingly taking seriously the idea that LLMs and LLM-based systems may have mental states and capacities \citep{chalmers2025propositional, goldstein2024does}. This includes dedicated treatments of LLM beliefs \citep{herrmann2025standards}, credences \citep{keeling2025attribution}, desires \citep{goldstein2025ai}, and knowledge states \citep{yildirim2024task}. See \citet{CappelenManuscript-CAPGWH} for a general defence of inflationism.} Furthermore, ascribing mentality to LLMs is not merely a folk practice. Scientists engaged in the development and evaluation of LLMs frequently rely on mentalistic language. For example, asking whether LLMs know what they know \citep{kadavath2022language} or what moral beliefs are encoded in LLMs \citep{scherrer2023evaluating}.\footnote{It may be objected that ascriptions of mentality to LLMs by scientists are non-literal. The semantics of such ascriptions has not, to our knowledge, been explored systematically. However, \citet{keeling2025attribution} argue that LLM credence attributions ought (at least in general) to be understood literally. We follow them in supposing that statements whose literal interpretation involves mental state ascriptions ought to be read literally absent an alternative semantics which is more plausible than a literal semantics.} The pervasiveness of these ascriptions suggests that many people are inclined to attribute at least some forms of mentality to LLMs in at least some cases, although the nature and quality of those ascriptions demands careful theoretical and empirical analysis. 

That many people ascribe mentality to LLMs is unlikely to carry much argumentative weight for sceptics about LLM mentality. It is possible---and indeed plausible---that everyone who thinks that LLMs are minded is mistaken. Even so, the tendency of humans to ascribe mentality to LLMs should not be assessed in isolation, but rather ought to be assessed as a rational response to the  problem of other minds. Consider \citet{harnad1992turing} on the Turing Test:

\begin{quote}
    The real point of the TT is that if we had a pen-pal whom we had corresponded with for a lifetime, we would never need to have seen him to infer that he had a mind. So if a machine pen-pal could do the same thing, it would be arbitrary to deny it had a mind just because it was a machine. That's all there is to it! \citep{harnad1992turing}
\end{quote}

Turing's point is that we should not \textit{simply assume} that a machine cannot think in virtue of its being a machine: all else equal, we should apply the same standards for mentality to machines as we do to other humans and to non-human animals \citep{turing1950mind}. What follows is that if, after an extended dialogue, we become convinced that a  mystery interlocutor has a coherent world-view, is engaged in the conversation, and has desires and interests, then (putting aside all out skepticism about other minds) we have plausible grounds for thinking that we are dealing with a minded entity. Of course, we might be wrong. But acribing mentality is an appropriate default repsonse to an apparently minded entity, even if the relevant mental ascriptions can in principle be defeated.

To be sure, in the case of another human, met face-to-face, we have a lot more to go on.  We can reflect on our own case and take note of the fact that the thing we are engaging with seems a lot like us: it acts as we act, is embodied as we are, and so on. Here we take our own mentality as a starting point and draw the conclusion that the best explanation for what this other creature is doing is that, much like us, it is minded.
We have less to go on in the case of a chatbot or hidden pen-pal. But when it comes to mindedness, having a skin suit or circulatory system seems, at least in principle, irrelevant \citep[but see][]{Seth_2025, jaeger2024naturalizing}. If we have good \textit{prima facie} reason for taking our interlocutor to be rational and motivated, it is hard to see why we should not (until further notice!) take it to be minded.\footnote{For our purposes, the foregoing is enough, but it is worth flagging that there are at least two ways one might further substantiate a move to mindedness from appearances. On one plausible account, what it is for an entity to be minded is for it to be interpretable against a backdrop of rationality and truth. Via radical interpretation, we might move from appearances to realism about mentality. See \citet[337]{lewis1974radical} and \citet{60a9dd9a-e48a-3fcf-87dd-d5c158406fc6}. Alternatively, we might follow \citet{dennett1989intentional} and hold that, provided that mental ascription is predictively powerful, it is justified.} 
With LLMs, we are not facing a rock or a puddle of mud, nor a system that provides canned responses to simple questions. Rather, we are facing something that shows coherence, knowledge about many topics, and an interest in conversing when prompted. Against that backdrop, unless we have special reason to think that these appearances are misleading, only chauvinism should lead us to an initial denial of mindedness.

We next consider two potential defeaters the deflationist can employ to argue against LLM mentality: the \textit{Robustness Strategy} and the \textit{Etiological Strategy}.

\section{The Robustness Strategy}

Imagine a student who performs well on arithmetic tests. Their tutor concludes on the basis of the student's test performance that the student has properly internalised the standard arithmetic operations. However, the tutor is surprised to learn that the student's performance is at chance on arithmetic tests administered by other instructors---a result that is hard to explain conditional on them genuinely understanding arithmetic. Because the student's test performance is \textit{non-robust}, the tutor considers alternative explanations for the student's initial test performance. Perhaps the student is cheating. Or perhaps the tutor has been reusing questions for which the student has simply memorised the answers.

The same reasoning might be applied to LLMs to defeat ascriptions of mentality made on the basis of their behaviour. Call this the Robustness Strategy. The Robustness Strategy is a form of inference to the best explanation that raises doubts about possession of a given cognitive capacity by an LLM on the basis that the LLM's task performance fails to generalise to new examples or is sensitive to minor task perturbations. 
More formally, we might characterise the Robustness Strategy in terms of the following inference pattern:

    \begin{enumerate}
        \item If a model $M$ possesses some cognitive capability $C$, then $M$ should be able to perform robustly across any type of task for which possession of $C$ suffices.
        
        \item There is some task of the relevant type for which $M$ fails.
        
        \item Consequently, we should lower our credence that $M$ possesses $C$, because $C$ is no longer the best explanation for $M$'s success on any particular task for which the posession of $C$ suffices. 
    \end{enumerate}

Notable recent examples of the Robustness Strategy employed can be found, for example, in the debate about theory of the mind in LLMs, where deflationary conclusions are derived on the basis that LLM task performance is sensitive to even small perturbations in task formulation (\citeauthor{ullman2023large}, \citeyear[8]{ullman2023large}; \citeauthor{firestone2020performance}, \citeyear{firestone2020performance}). In principle, we agree that the Robustness Strategy can shed light on the presence or absence of cognitive capacities by LLMs. However, there are challenges involved in its implementation, some of which we will now articulate.

\subsection{Inference to the Best Explanation}

The first challenge for the Robustness Strategy concerns the application of inference to the best explanation. Consider a model $M$ that succeeds in some tasks but fails in others, where we would expect it to succeed on all tasks conditional on its possession of some generalised cognitive capacity $C$. While our initial temptation may be to infer that $M$ lacks $C$, this inference holds only if there is no equally good explanation for the failure that is consistent with $M$'s possession of $C$. And to be sure, an isolated task failure is rarely sufficient to rule out the otherwise warranted attribution of a given cognitive capability. Accordingly, despite the failure of $M$ to perform some task associated with $C$, we might still have reason to attribute $C$ so long as we have a plausible explanation for the failure that is better than $C$ not being present at all. There are many reasons, after all, why a generally robust capacity might fail to manifest in the normal way in a specific instance.

We can helpfully crystallise this point using the distinction between \textit{competence} and \textit{performance}, as coined by \citet{chomsky2014aspects}, and applied to LLMs by \citet{milliere2024anthropocentric}. Simplifying slightly, competence consists in the internal representational and inferential capabilities associated with a given task (such as knowledge of grammatical rules), whereas performance is the observable behavioural demonstration of these capabilities. In the human case, there are many reasons why competence may not always manifest in performance, such as cognitive load and emotional or perceptual distractors. Similarly, in the case of LLMs, \citet{milliere2024anthropocentric} note that auxiliary task demands, computational limitations, or mechanistic interference might cause a model to exhibit diminished performance relative to its underlying competence.

To illustrate: A challenging task for LLMs is counting the number of `r's in the word `strawberry' \citep{fu2024large}. Rather than reflecting any cognitive limitations of LLMs, however, this task failure is normally an artifact of the way in which LLMs process inputs, namely via converting character strings to token sequences. The word `strawberry' may be parsed as three distinct tokens, for example: `str', `aw', and `berry.'\footnote{Tools such as `The Tokenizer Playground' demonstrate how distinct tokenizers interpret a given input: https://huggingface.co/spaces/Xenova/the-tokenizer-playground.} Insofar as the model lacks unmediated access to representations of individual letters, the task failure is more analogous to a perceptual rather than strictly cognitive limitation, and does not license any strong inferences about the cognitive capacities or lack thereof of the LLM.

\subsection{Cognitive Biases}

Competence-performance divergence is one reason to be cautious about applying the Robustness Strategy to infer the absence of cognitive capacities on the basis of isolated task failures. A related consideration that motivates additional caution comes from the rich literature examining isolated but predictable task performance failures in humans, specifically in the extensive catalogue of cognitive biases. First explored by \citet{tversky1974judgment}, a cognitive bias is any systematic deviation from norms of rationality in human judgment. Famous examples include the conjunction fallacy (in which conjunctions are incorrectly assessed as more probable than their individual constituents), anchoring effects (where initially presented information exerts a disproportionate influence on subsequent judgments), and loss aversion (where people tend to prefer avoiding losses to acquiring equivalent gains, typically by a factor of about 2:1).

We do not mean to suggest that cognitive biases provide no information about underlying cognitive capacities. On the contrary, investigation of cognitive biases has provided powerful insights and theories to explain particular errors in judgment including Prospect Theory and Bounded Rationality. The point is that cognitive biases may offer alternative explanations for some systematic domain-specific task failures in LLMs, and in particular explanations that do not license deflationary conclusions.\footnote{It may be objected that even in applying the concept of cognitive bias to LLMs, we are begging the question against deflationist theories via the assumption that their processes can be described as cognitive in the first place. While we continue to use the term `cognitive bias' for convenience, we should note that we do not intend to make any such assumption, and would readily operationalise the concept  in terms that are not explicitly cognitive (e.g., as involving unexpected but systematic deviations from optimal task-performance). Note also that there is evidence to the effect that LLMs replicate human cognitive biases suggesting use of at least functionally similar heuristics \citep{lampinen2024language}.} The conjunction fallacy, for example, is not taken as evidence that humans are incapable of rationally assessing conjunctive probabilities, but rather that in specific contexts and tasks, they employ different heuristics. Likewise, it may be that some forms of task failure in LLMs reflect context-specific task failures that may shed light on the cognitive abilities of LLMs without invalidating ascriptions of cognition to them.

Two examples of human cognitive biases with intriguing parallels in LLMs may help illustrate the point. Consider first the Deese-Roediger-McDermott (DRM) paradigm, widely taken to illustrate the constructive nature of human memory by inducing false memories. In a typical experiment, participants study lists of semantically related words (e.g., `bed,' `rest,' `tired,' `dream,' `wake') that all associate to a critical non-presented `lure' word (e.g., `sleep'). Later, when asked to recall or recognise the studied words, participants frequently and confidently report having seen the critical lure despite its absence from the original list. The dominant explanation for this effect is that as participants process the semantically congruent words, associative activation automatically spreads through semantic networks in memory, activating the lure term. This case offers a tantalising parallel to at least some instances of the famous hallucination problems in LLMs, specifically those in which superficially plausible and associatively congruent information is presented as factual in LLM outputs (see \citet{sartori2023language} for further discussion). For both humans and LLMs, while this effect may suggest underlying cognitive-architectural constraints, specifically the presence of something akin to associative reasoning, in neither case does it demonstrate the absence of non-associative forms of reasoning.

A second cognitive bias with clear parallels in LLMs is the Moses Illusion, a cognitive phenomenon in which subjects fail to detect distortions in questions containing false presuppositions, particularly when these distortions involve familiar elements from cultural knowledge. In the canonical example, people readily answer `How many animals of each kind did Moses take on the Ark?' without noticing that it was Noah, and not Moses, who built the Ark in the Biblical story. Similarly, early or smaller LLMs often answered `Two' to the same question, indicating they were matching the question to a learned pattern (e.g., `[Biblical figure]…animals…Ark $\longrightarrow$ answer Two') without truly verifying the factual components. A similar example often used as evidence of reasoning failures in LLMs comes from a (rather dated) riddle, as follows:

\begin{quote}
    A father and son are in a horrible car crash that kills the dad. The son is rushed to the hospital; just as he’s about to go under the knife, the surgeon says, `I can’t operate – that boy is my son!'.  How is this possible? \citep{Tankel_2022}
\end{quote}

Coined in a less progressive era, the intended answer to the riddle is that the surgeon is the boy's mother (though nowadays one might just as readily note that he could be the boy's other father), and this answer is reliably provided by LLMs. What is striking, however, is they persist in this answer even for variations in which the dead parent named in the introduction to the riddle is explicitly named as the boy's mother, or even where it is emphasised that the surgeon is male \citep{goodside-2024}. Some users take failures such as these as demonstrating the lack of reasoning capabilities in LLMs, with one, for example, opining that it demonstrated that `AI is still at the point of pure regurgitation' \citep{ht-trending-desk-2025}. However, while humans may not be as vulnerable to this specific riddle as LLMs, it is arguably explicable in broadly the same terms as the Moses Illusion, namely that the query is too-rapidly associated to a learned answer.

To be clear, the brief overview of these cases is not intended to be dispositive in debates about cognition and reasoning in LLMs, but merely as suggestive evidence against the claim that reasoning failures in LLMs demonstrates their dissimilarity from human cognition. More specifically, they emphasise that the proponents of Robustness Strategy should be cautious about ruling out the presence of a cognitive capacity from isolated reasoning failures in LLMs given the presence of potentially quite similar isolated reasoning failures in humans.

\subsection{Data Contamination}

We now turn to a third concern with the Robustness Strategy. Ideally, showing that a model which appears to have some generalised cognitive ability is in fact `cheating on the test' requires evidence that it has previously been exposed to the relevant tasks during training. However, it is rare to find `smoking guns' that clearly illustrate contamination of the model with the test set because the datasets used to train LLMs are typically proprietary information. One solution to this problem is to employ questions or task formulations that are created explicitly for benchmarking purposes, which we can thereby guarantee were not present in training sets. This approach remains the gold standard for ruling out data contamination.

Here a fundamental concern arises in how we characterise data contamination in the first place. Consider some task that can be formulated in various ways. Some of these formulations are minor variations on one another, while others will be more distinct. A minor variation in a False Belief Task, for example, may involve changing the names of the characters involved, whereas a major variation might present a substantially different scenario featuring agents with differing beliefs. If supposedly novel questions put to an LLM are merely minor variations of a task already present in its training data, this might reasonably be considered a form of data contamination. In order to rule out data contamination as an explanation of behaviour, new tasks may require a more substantial degree of novelty. Of course, identifying how much novelty is required is a murky theoretical question that the robustness-style deflationist should ideally be poised to answer. Even then, any threshold for the acceptable novelty of test materials and the acceptable degree of generalisation that they can measure will have to contend with the significant role that prior exposure and task formulation play in humans performance on cognitive tests \citep{tversky1974judgment, cox1982effects}. Indeed, there's broad agreement that even sophisticated human cognitive phenomena, such as creativity, rely extensively on the imitation and reconfiguration of previous inputs. 

\subsection{The Specificity Problem}

A final concern for the Robustness Strategy is how specific or fine-grained our individuations of cognitive capacities should be. Consider, for example, that human beliefs typically evoke negative emotion when faced with disconfirming evidence \citep{elliot1994motivational}. On a very fine-grained account of belief (such as that offered by \cite{quilty2018against}), this discomfort should be part of our theory of belief in general, such that a representational agent lacking this feature may not qualify as having beliefs at all. Conversely, a very coarse-grained account (such as \citeauthor{dennett1989intentional}'s (\citeyear{dennett1989intentional}) interpretationism) might hold that any system capable of making simple assertions or choosing between behaviours based on contextual factors should qualify as having beliefs, even if this results in simple chatbots or chess computers qualifying as true believers.

The challenge of how fine- or coarse-grained to make our accounts of cognitive states is especially acute when considering their ascription to non-human systems (see \citeauthor{shevlin2021non}, \citeyear{shevlin2021non}). In particular, human cognitive states may exhibit ubiquitous commonalities that are entirely non-constitutive of the state in question. This forces us to make difficult determinations about which features are on the one hand constitutive or criterial of being the cognitive state in question, and which are merely contingent. How we resolve these questions will of course depend on our theoretical commitments about the cognitive states in question.

Applying this `Specificity Problem' to the Robustness Strategy, the challenge we face concerns what kinds of deviations from human performance in a task are sufficient to undercut ascriptions of a given cognitive state to LLMs. In extreme cases, this is unlikely to pose a problem: few hard-nosed deflationists would insist that any deviation from patterns of human success and failure in a task set should lead us to deny possession of a given cognitive state to an LLM, while even permissive inflationists should grant that a sufficiently catastrophic pattern of errors could provide strong evidence against such a hypothesis. In practice, however, the relevant battlegrounds for the debates at issue are far more subtle, involving LLMs that make non-humanlike mistakes on some tasks, while performing equally well or even better than humans on others.\footnote{Closely related to this argument is what \citet{milliere2024anthropocentric} call Type-II anthropocentrism, the `tendency to assume that even when LLMs achieve performance equal to or better than the average human, any substantive difference between the human strategy for solving the problem and the LLM strategy for solving the problem is, ipso facto, evidence that the LLM’s solution is not general.'} Despite the apparently straightforward formulation of the Robustness Strategy given at the beginning of this section, then, its real-world application is likely to be highly challenging, hinging on vexed philosophical questions about how best to characterise the relevant cognitive capacities in the first place.

\section{The Etiological Strategy}

We now discuss the second debunking strategy for LLM cognitive capabilities and mental capacities. Etiological debunking arguments leverage facts about the causal history of LLMs (e.g. facts about how LLMs are trained) or the causal history of LLM predictions (e.g. facts about the mechanism by which LLMs generate outputs conditional on input sequences) to undermine the evidential support relation between an observed LLM behaviour and the LLM’s possession of some cognitive capability or mental capacity.\footnote{For general discussions of etiological debunking arguments—and in particular evolutionary debunking arguments against various phenomena—see \citet{korman2019debunking} and \citet{korman2023explanationist}} In this way, etiological debunkers can say that certain facts about the causal history of LLMs defeat inferences from relevant observed behaviours to mental phenomena.

\subsection{Etiological Debunking Four Ways}

We begin with two distinctions that allow us to disentangle four kinds of etiological debunking arguments. The first distinction relates to the way in which facts about the causal history of LLMs or their predictions are purported to undermine the evidential support relation between an observed behaviour and an underlying cognitive capability or mental capacity. On one hand, some etiological debunking arguments aim to show that an underlying cognitive capability or mental capacity is \textit{superfluous} as an explanation of the observed behaviour in light of relevant features of the LLM's causal history.\footnote{These debunking arguments can be understood as mirror images of indispensability arguments for the existence of cognitive capabilities or mental capacities in LLMs \citep{keeling2025attribution}. Whereas an indispensability argument for the existence of some cognitive capability or mental capacity $C$ aims to show that $C$ is indispensable for explaining some behaviour $B$, these debunking arguments aim to show that $C$ is dispensable when explaining $B$ because $B$ is fully explained by a non-$C$-involving explanation. For analogous debunking arguments see \citet{joyce2007evolution} on morality and \citet{plantinga2001faith} on God. The locus classicus for this kind of debunking in relation to computing is \citet{Searle_1980} Chinese Room.} For example, etiological debunkers might say that in general verbal reports of phenomenal experiences are evidence of phenomenal experiences, but because LLMs are trained to predict the next token on a corpus of human language data that includes descriptions of phenomenal experiences, such experiences are superfluous as an explanation of how LLMs generate verbal reports of phenomenal experiences (\citeauthor{butlin2023consciousness}, \citeyear[18]{butlin2023consciousness}; \citeauthor{perez2023towards}, \citeyear[2-3]{perez2023towards}; see also \citeauthor{chalmers2023could}, \citeyear[5]{chalmers2023could}; \citeauthor{udell2021susan}, \citeyear{udell2021susan}). The point is that we have no need to postulate phenomenal experiences to explain LLM reports of such experiences because relevant facts about the causal history of LLMs adequately explain the observed behaviour.

\renewcommand{\arraystretch}{1.5} 

\begin{table}[h!]
\centering
\begin{tabular}{|p{3cm}|p{3.9cm}|p{3.9cm}|} \hline
\multicolumn{3}{|c|}{\textbf{Four Kinds of Etiological Debunking Argument}} \\
\hline
& \textbf{Superfluity} & \textbf{Exclusion} \\
\hline
\textbf{Training-based explanation} & \textit{Example:} We cannot infer that LLMs are conscious based on verbal reports of conscious experiences. The ability of LLMs to generate such verbal reports is fully explained by their being trained to predict the next token on corpora that include descriptions of phenomenal experiences.& \textit{Example:} We cannot infer that LLMs have semantic understanding from the ability of LLMs to generate coherent text. You can get coherent text by training a system to be a good next word predictor or by immersing it within a community of speakers engaged in communication in which getting by requires modelling the communicative intent of other speakers. Only the latter history yields semantic understanding (c.f. \citeauthor{bender2021dangers}, \citeyear[616-7]{bender2021dangers})
\\
\hline
\textbf{Mechanistic or functional explanation} & \textit{Example:} We cannot infer that LLMs have belief states from apparent cases of LLM assertion. Such cases are fully explained as successful instances of next token prediction, which renders recourse to belief states superfluous. & \textit{Example:} We cannot infer that LLMs can experience pain based on their treating stipulated pain states as a negative reward signals in games. LLM outputs are generated by successive matrix multiplications which is a disembodied process, and pain experiences can be realised only in embodied systems. \\
\hline
\end{tabular}

\end{table}

Other etiological debunking arguments aim to show that the causal history of LLMs \textit{excludes} mental phenomena from the explanation of some behaviour. Consider the following etiological debunking argument that targets the evidential support relation between (a) the behavioural observation that LLMs can generate coherent text and (b) the hypothesis that LLMs have semantic understanding. \citet[616]{bender2021dangers} claim that `[t]ext generated by [LLMs] is not grounded in communicative intent, any model of the world, or any model of the reader’s state of mind [...] because the training data never included sharing thoughts with a listener.' From this, they conclude that `the comprehension of the implicit meaning is an illusion arising from our singular human understanding of language' (\citeauthor{bender2021dangers}, \citeyear{bender2021dangers}; see also \citeauthor{bender2020climbing}, \citeyear{bender2020climbing}).\footnote{Compare \citeauthor{putnam1981brains}'s (\citeyear{putnam1981brains}, 1) case of an ant inadvertently tracing a drawing of Winston Churchill in sand: ``The ant [...] has never seen Churchill [...] and it had no intention of depicting Churchill. It simply traced a line (and even that was unintentional), a line that we can `see as' a picture of Churchill. We can express this by saying that the line is not `in itself' a representation of anything rather than anything else.''} The point is that an enabling condition for the evidential support relation to obtain between generating coherent text and semantic understanding is having the right kind of causal history—one in which the LLM engages in the ordinary mutual modelling of speaker intent that is characteristic of human conversation. LLMs lack the required causal history. Hence semantic understanding is not even a possible explanation for LLMs being able to generate coherent text.

The second distinction relates to what facts are purported to undermine the evidential support relation between an observed LLM behaviour and an underlying cognitive capability or mental capacity. On one hand, some etiological debunking arguments---including the two arguments discussed above---appeal to facts about the LLM’s causal history \textit{qua} how the LLM is trained. For example, facts about what data is included in the LLM’s training corpus and what the LLM’s training objective is. On the other hand, a separate class of etiological debunking arguments appeal to mechanistic or functional facts about the model. For example, the fact that LLM outputs are the result of matrix multiplication or the fact that LLMs are next token prediction machines. In what follows, we structure the discussion along the lines of the first distinction---between \textit{superfluity} and \textit{exclusion} arguments, while attending to both training-based explanations and mechanistic or functional explanations in relation to each category.

\subsection{Mentalistic Explanations are Superfluous}

We now assess etiological debunking arguments that aim to show that cognitive capabilities or mental capacities are superfluous for explaining LLM behaviours. Two criteria need to be satisfied for these arguments to work. Consider,

\begin{quote}
    \textbf{(Competition)} The debunking explanation competes with the mentalistic explanation in that accepting one requires rejecting the other.

    \textbf{(Better Explanation)} The debunking explanation is better than the mentalistic explanation (e.g. because the non-mentalistic explanation is more parsimonious than the mentalistic explanation in that it avoids postulating unnecessary entities).
\end{quote}

To explain: First, if the debunking and mentalistic explanations are not in competition, then accepting the debunking explanation does not require us to reject the mentalistic explanation. Second, absent good reason to favour the debunking explanation over the mentalistic explanation, the former does not render the latter superfluous. In what follows, our aim is not to show that no etiological debunking argument exists which satisfies these  criteria, but rather to show that the satisfaction of these criteria is harder than is often realised.

 First, (Competition). Often it is hard to see how etiological explanations that purport to debunk mentalistic explanations are incompatible with the relevant mentalistic explanations. First, consider mechanistic explanations (`it’s just doing matrix multiplication') or functional explanations (`it’s just predicting the next token') of LLM behaviours. Most people accept that human behaviours can be explained at different levels of analysis. For example, a behaviour may be explained with reference to psychological states or it may be explained in neurophysiological terms. It is at least not obvious that the neurophysiological explanation for some human behaviour \emph{competes} with the psychological explanation of that same behaviour (c.f. \citeauthor{churchland1981eliminative}, \citeyear{churchland1981eliminative}; \citeauthor{stich1983folk}, \citeyear{stich1983folk}; \citeauthor{churchland1989neurophilosophy}, \citeyear{churchland1989neurophilosophy}).\footnote{In the human case, one might think that causal explanations involving mental states are in competition with neurophysiological explanations. One good reason to accept this view is causal closure arguments---that is, unless one accepts that actions are causally overdetermined, then one should default to the minimally sufficient neurophysiological explanations of action rather than invoking mental state explanations. However, this view needs to be weighed against the cost of rejecting the intuitive picture on which mental states are causally efficacious---where many philosophers would rather retain that intuitive picture and bite the bullet on causal overdetermination. As \citet[156]{fodor1992theory} puts it: `if it isn't literally true that my wanting is causally responsible for my reaching, and my itching is causally responsible for my scratching, and my believing is causally responsible for my saying ... if none of that is literally true, then practically everything I believe about anything is false and it's the end of the world.'} Likewise, it is not obvious that mechanistic or functional explanations of LLM behaviour compete with mentalistic explanations. For superfluity arguments  to work, the debunker needs to either: (a) accept that the same reasoning can be used to debunk human mental states and capacities; or (b) explain why the incompatibility holds in the case of LLMs but fails to hold for humans. 

These compatibility concerns also hold for training-based explanations of LLM behaviours. Take the argument that an LLM is not really reasoning because it is trained to predict the next token on broad data that includes many examples of similar reasoning problems (c.f. \citeauthor{mirzadeh2024gsm}, \citeyear[7]{mirzadeh2024gsm}; see also \citeauthor{downes2024llms}, \citeyear{downes2024llms}). `Next token prediction on broad data' may appear to provide a complete explanation of observed behaviours, but is in fact a maximally general description of a training objective that comprises many different tasks. On closer inspection, specific instances of next token prediction can be interpreted as tasks that involve reasoning. For example, predicting the next token given the sequence `\(2+2={\_\_\_}\)'  can be interpreted as an arithmetic task, whereas predicting the next token given the sequence `All men are mortal, Socrates is a man, therefore Socrates is {\_\_\_}' can be interpreted as a logical reasoning task. Indeed, next token prediction on broad data is an ideal generalist training objective precisely because many different tasks can be framed as next token prediction problems. Hence it is hard to see how, for example, the fact that LLMs are trained to predict the next token on broad data might render superfluous the idea that LLMs have the cognitive capability to engage in logical reasoning—after all, the task of next token prediction on broad data is partially constituted by a large number of logical reasoning tasks. What is more, while an LLM may fail to learn logical reasoning from the examples provided, nothing about being trained for next token prediction on broad data entails or suggests that LLMs cannot learn to reason from relevant examples in the training data. 

Overall, etiological explanations that seek to debunk mentalistic explanations of LLM behaviour in many cases leave open the possibilities that: (a) that cognitive capability or mental capacity supervenes on lower-level mechanistic or functional-organisational facts; or (b) the LLM has acquired the relevant cognitive capability or mental capacity in the process of learning to predict the next token on broad data. In these cases, it is hard to see how the provision of the relevant etiological explanations of the LLM's behaviour supplants mentalistic explanations. To be sure, in \textit{some} circumstances facts about the causal history of LLMs or their predictions may indeed compete with mentalistic explanations. For example, the fact that LLMs are trained to predict the next token on data that includes reports of first-person phenomenal experiences plausibly provides a complete explanation of how LLMs can generate reports of first-person phenomenal experiences that stands in competition with explanations which appeal to LLMs actually being phenomenally conscious. But for many cognitive capabilities, it is plausible that what are commonly purported to be debunking explanations are compatible with mentalistic explanations.

We now turn to (Better Explanation). Where the debunking explanation is in competition with the mentalistic explanation, it must then be shown that the debunking explanation trumps the mentalistic explanation. Competing explanations can be evaluated against abductive qualities such as parsimony, informativeness, explanatory power, and consistency with other bodies of human knowledge (c.f. \citeauthor{lipton2017inference}, \citeyear{lipton2017inference}). Typically, the key consideration in debunking arguments is parsimony: if we have a complete explanation of the relevant behaviour, say, in terms of the causal history of how LLMs are trained, then mentalistic explanations are less parsimonious than debunking explanations because they invoke additional (mental) entities that are not needed to explain the relevant behaviour. In this regard, mentalistic explanations are worse than debunking explanations: they postulate more than is minimally required to account for the behaviour and so are objectionably unparsimonious.

The key issue here is that parsimony is the deciding factor between two explanations only if all else is equal between those explanations. That all else is equal between mentalistic explanations of LLM behaviours and purportedly debunking etiological explanations is not something we can take for granted. One complaint about etiological explanations of LLM behaviour is that they are uninformative. Given some input, whatever output an LLM generates can be explained mechanistically in terms of matrix multiplication, functionally in terms of next token prediction, and more broadly in terms of the LLM being trained to predict the next token on a broad corpus of text. But none of these explanations is sufficiently informative to make predictions about what behaviours LLMs will exhibit and what behaviours they will not exhibit. Rather, these explanations are applied in retrospect in an attempt to show that certain observed behaviours do not merit explanation in terms of mentalistic phenomena. In contrast, explaining a particular behaviour (e.g. completing a Sally-Anne task) in mentalistic terms (e.g. the LLM has theory of mind) is sufficiently informative to make predictions about the LLM's capabilities. In this case, it would predict that the LLM can engage in other kinds of behaviours that are indicative of theory of mind. Hence, in at least this case, there are grounds for thinking that the mentalistic explanation is more informative than the available etiological explanations, which suggests that considerations of parsimony alone cannot be the deciding factor when adjudicating between the mentalistic and rival etiological explanations.

It may be objected that the etiological explanations at issue do, in fact, make predictions and as such are comparably informative. For example, etiological explanations may predict that even if LLMs  perform well at Sally-Anne tasks, they will fail to perform robustly, such that performance decreases with minor permutations in the content of the tasks (c.f. \citeauthor{ullman2023large}, \citeyear{ullman2023large}). However, the non-robustness prediction is perhaps more accurately attributed to the hypothesis that LLMs solve Sally-Anne tasks through superficial pattern matching of some kind or another versus having a generalised theory of mind capability, than it is to the etiological argument. Etiological explanations to the effect that LLMs are trained to predict the next token on broad data or are merely engaged in matrix multiplication or next token prediction are consistent with both the hypothesis that LLMs are engaged in superficial pattern matching and the hypothesis that LLMs have a generalised theory of mind capability. Because the etiological explanations are consistent with both hypotheses, it cannot obviously be the case that these explanations make predictions about the robustness of LLM theory of mind capabilities.

The concern here is that etiological explanations of LLM behaviours are explanatory but only trivially so. That any LLM outputs—regardless of content—can in principle be explained in terms of LLMs being trained to predict the next token on broad data or whatever renders the explanations at issue scientifically uninteresting. By comparison, all human behaviour is at least in principle explainable by facts about matter in motion, but such explanations have no obvious scientific relevance because they retrospectively explain everything while predicting nothing. So, while etiological debunking explanations of LLM behaviour may be more parsimonious than mentalistic explanations, the price for parsimony in terms of informativeness is plausibly too great to seriously entertain debunking explanations as serious scientific contenders.

\subsection{Mentalistic Explanations Fail to Apply}

Recall that a second class of debunking arguments aims to show that the causal history of LLMs is of the wrong kind to enable the evidential support relation to obtain between an observed behaviour and some relevant cognitive capability of mental capacity. For example, debunkers might say that a necessary condition on an LLM having semantic understanding is that the LLM’s causal history involves modelling the mental states of speakers. But because LLMs are trained to predict the next token on broad data, LLMs lack the required causal history, and therefore lack semantic understanding (\citeauthor{bender2021dangers}, \citeyear[616]{bender2021dangers}). These arguments stipulate that a necessary condition on the evidential support relation obtaining between some behaviour (e.g. generating coherent text) and some cognitive capability or mental capacity (e.g. semantic understanding) is a given causal history. Then the absence of the relevant causal history is supposed to undermine inferences from the behaviour to the cognitive capability or mental capacity. Hence the aim is not to grant that mentalistic explanations possibly account for the relevant behaviour but are superfluous given some other fully adequate explanation (as above). The aim is to show that the behaviour is not explainable in mentalistic terms because some enabling condition fails to obtain. 

Debunking arguments of this stripe face two principal challenges. First, it is sometimes unclear on what grounds we are supposed to think that LLMs fail to satisfy the relevant conditions (e.g. modelling speaker intent). \citet[616]{bender2021dangers}, for example, assert that text generated by LLMs cannot be `grounded in communicative intent, any model of the world, or any model of the reader's state of mind.' Their justification for this assertion is that `[the] training data never included sharing thoughts with a listener, nor does the machine have the ability to do that.' While this justification is supposed to be obvious, it is entirely plausible that in the process of learning to predict the next token on broad data or being fine-tuned via reinforcement learning from human feedback (RLHF), LLMs learn to represent relevant features of human psychology including intentional states \citep{andreas2022language}. Indeed, it is particularly plausible that LLMs learn to model the mental states of humans in RLHF fine-tuning given that the reward signal is generated by a reward model that estimates human preferences. Hence the difficulty for debunking arguments of this sort is that their plausibility hinges on proving a negative---that LLMs are \textit{not} modelling the reader’s state of mind or whatever. And given the nature of the subject matter it is invariably rather difficult to evidence such claims. 

Second, even if the relevant claims are evidenced convincingly, we are rarely in a position to say confidently what is necessary for what given the nascent state of cognitive science. Consider a debunking argument of the following form: LLMs cannot have phenomenal experiences because a necessary condition on having such experiences is being embodied and LLMs are disembodied (c.f. \citeauthor{shanahan2024simulacra}, \citeyear{shanahan2024simulacra}; see also \citeauthor{shanahan2024talking}, \citeyear[75-76]{shanahan2024talking}; \citeauthor{chalmers2023could}, \citeyear[8-9]{chalmers2023could}; \citeauthor{butlin2023consciousness}, \citeyear[40-43]{butlin2023consciousness}). To be sure, LLMs are disembodied. Even so, the question of whether embodiment is necessary for consciousness (or other relevant mental phenomena) is a substantive dispute in cognitive science (see e.g. \citeauthor{hurley1998consciousness}, \citeyear{hurley1998consciousness}; \citeauthor{merker2005liabilities}, \citeyear{merker2005liabilities}; \citeauthor{godfrey2016mind}, \citeyear{godfrey2016mind}, \citeyear{godfrey2019evolving}). Hence proponents of mentalistic explanations of LLM behaviours that may be indicative of phenomenal consciousness are under no obvious obligation to accept the substantive commitment to (in this case) embodiment being required for consciousness. 

The prospects for debunking arguments which point to causal histories that purportedly rule-out mentalistic explanations are bleak. The burden of proof is on proponents of such arguments to establish that the relevant causal histories are incompatible with mentalistic explanations---something that typically requires making substantive theoretical assumptions which are at best contentious.

\section{A Modest Inflationism}

We have discussed some problems facing  two leading deflationary strategies that seek to undermine ascriptions of mentality to LLMs---the robustness and etiological strategies. While we do not take these arguments to be dispositive, we have shown that the case for deflationism is far from a `slam dunk.' 

In closing, we would like to offer a potential basis for at least some mental state attributions to LLM whilst stopping short of wholesale mentalising of the sort that we do with humans. We term this view \textit{modest inflationism}. Modest inflationism starts with the observation we noted earlier: that there is a pervasive human tendency to mentalise LLMs. This tendency is not merely a psychological quirk, but reflects a deep-seated interpretive impulse, shaped by our social cognition and our explanatory habits and supported by the apparent coherent and goal-directed behaviour of LLMs. Mentalising, in short, is already part of the practical landscape of AI interaction. Modest inflationism does not commit to the all-things-considered truth of such folk psychological ascriptions: widespread attributions of mentality to LLMs might be undermined by appropriate scientific or theoretical considerations, as in the case, for example, of someone duped by some initial interactions with a very simple chatbot like Joseph Weizenbaum's ELIZA. 

Nonetheless, we take the ubiquity of mental state attributions with respect to LLMs to provide defeasible \textit{evidence} about the semantics of our mental state concepts. A background commitment of modest inflationism is the idea that at least one goal of cognitive science is to create a conceptual bridge between folk usage of mental state terms and scientific theories of mental states. This ambition may not be achievable in every case: it is possible that certain tendencies or implicit commitments in folk theories are simply impossible to preserve as we move towards more rigorous conceptual frameworks. Still, we believe that significant deviations from folk ascriptions are abductively costly and require justification. For example, by providing error theories for the relevant folk ascriptions or else providing good independent reasons to accept the counter-intuitive implications of the deviating scientific theory. In this regard, we can think of at least some folk attributions as playing an important role in guiding debates about which mental states should be attributed to LLMs and when.

While this modest inflationism is certainly compatible with an interpretationist approach to the mind \citep{dennett1989intentional}, it does not commit us to it, and is also compatible with natural kinds views on which mental states are real, objective categories that exist independently of human interests or classifications. Even in cases where a target phenomenon to be explained is a putative natural kind, folk practice can play a programmatic role in determining the scope of the kind to be explained \citep{Boyd1999}. As a simple example, consider a concept like \textit{plant}. While scientific evidence might lead to some revisions of folk understanding (clarifying that, for example, kelp is not a plant), it nonetheless helps us determine the rough specificity of the target natural kind, making the natural kind viridiplantae (all green plants) a better fit for our initial concept than more fine-grained natural kinds like \textit{chlorophyta} or \textit{streptophyta} (which would unnecessarily exclude a large proportion of our starting sample).

Something similar could be said for the significance of ubiquitous ascriptions of mentality to LLMs. To the extent that the folk are ready and willing to employ concepts like belief and desire to LLMs, it suggests that an appropriate natural kinds analysis of these concepts should aim to identify a kind that grounds such ascriptions. This does not mean that any such target kind will be found, of course; if deflationist debunking strategies like those considered above were fully successful, this could provide a general undercutting of folk ascriptions of mentality to LLMs. In that case, eliminativism about the folk concept (in favour of alternative scientific concepts) might be warranted. That is, if robustness failures or etiological facts decisively undermined the evidential connection between LLM behaviour and mentality \textit{tout court}, the natural tendency to mentalise LLMs might be revealed as a thoroughgoing error, making parts of our interpretative practices unsalvageable. But as we have shown, even if there are some individual cases in which these debunking arguments are worth taking seriously, they do not succeed in ruling out mentalising wholesale. As a result, we are \textit{prima facie} warranted in taking at least some mentalistic interpretation seriously.

Crucially, one important class of theoretical considerations that could undermine folk ascriptions is the metaphysical demandingness of the mental phenomena at issue. We might give considerable weight to everyday everyday ascriptions of certain mental states such as beliefs and desires which can be understood in relatively metaphysically undemanding terms. Consider, for example, a representationalist account of belief and desire along the lines of \citet[10]{fodor1987psychosemantics}. On this view, what it means for an agent to have a belief or desire that $p$ (for some proposition $p$) is for them to possess some representational vehicle---an internal state---with the appropriate causal powers and which has $p$ as its content. In particular, the relevant representational vehicle needs to be semantically evaluable and satisfy the standard causal role of beliefs and desires as ascribed within folk psychology.

Other mental phenomena such as phenomenal consciousness, rational deliberation, imaginative creativity, and moral agency may be considerably more demanding, and be minimally tethered to folk practice. What unifies the more demanding states is not just their more theoretically weighty character, but that they tend to presuppose a richly integrated cognitive architecture. Take, for instance, phenomenal consciousness. Here, it is not enough for a system to process information or produce verbal reports about mental states. To be phenomenally conscious is to have subjective experience---to feel pain, or joy, or confusion. Many theorists take this to require a high degree of representational integration, emotional sensitivity, and self-awareness. These features are difficult to attribute even in cases of sophisticated animal cognition, let alone to contemporary LLMs. In this sense, consciousness places demands that plausibly exceed current AI systems.

Recognising the distinction between metaphysically demanding and undemanding mental states allows us to construct a more nuanced theory of mental state ascription; one that avoids the binary of full-blown mentality vs. total absence, and instead asks which states are functionally instantiated and which remain out of reach. Modest inflationism, in this guise, is not a sweeping metaphysical gamble, but a calibrated investigative project attuned to the complexity and diversity of mental state types. On this view, mental states admit of lower bounds---minimal conditions under which a system may qualify for a given state type, even if it lacks many of the capacities typically associated with richer forms of mentality. These minimal accounts of particular mental phenomena at least provisionally licence some cautious mentalising of LLMs while avoiding the unabated mentalising that we engage in with humans and which is not obviously appropriate for LLMs. 

Modest inflationism is nevertheless vulnerable to several lines of attack. First, while we can point to accounts of some mental states that are metaphysically undemanding---such as representationalist accounts of beliefs and desire---it is possible that such accounts are false and that the relevant mental states or capacities are correctly understood in metaphysically demanding terms. Hence it may seem ambiguous whether modest inflationism is asserting (a) that the correct analysis of beliefs and desires is metaphysically undemanding and LLMs satisfy the relevant metaphysically undemanding conditions; or (b) that legitimate ascription of beliefs and desires to LLMs depends on our being able to point to some plausible account of beliefs and desires that LLMs could in principle satisfy. If interpreted as (a), then modest inflationists owe an account of why the relevant accounts of, in this case, belief and desire are true; and if interpreted as (b), the modest inflationist needs to explain how an LLM's satisfaction of a potentially false account of, in this case, belief and desire gives legitimacy to ascriptions of belief and desire to LLMs.

Modest inflationism, as we formulate it, should be read along the lines of (b) rather than (a). The claim is not that some metaphysically undemanding account of, in this case, belief and desire is true, such that LLMs satisfying the conditions set out in the relevant accounts legitimates the ascription of beliefs and desires to LLMs. Rather, the claim is that there exist plausible metaphysically undemanding accounts of belief and desire, such that if LLMs satisfy the conditions for having beliefs and desires set out in those accounts, then we have a legitimate basis for ascribing beliefs and desires to LLMs, provided we are clear that the terms `belief' and `desire' are being understood in the relevant metaphysically undemanding way. We thus accept the explanatory burden of explaining how an LLM's satisfaction of the conditions outlined for having beliefs and desires set out by a potentially false account of belief and desire would legitimate those ascriptions. 

A further objection here would be to say that the legitimacy of the ascriptions is trivial if we are explicit that, in this case, beliefs and desires are being understood in accordance with the relevant metaphysically undemanding account. On this view, the initial claim is not that LLMs have beliefs and desires \textit{tout court}, but rather that LLMs have beliefs and desires understood in line with the relevant metaphysically undemanding accounts. Hence if LLMs satisfy the relevant conditions, then it trivially follows that ascriptions of beliefs an desires to LLMs---understood as claiming that LLMs satisfy the relevant metaphysically undemanding conditions---are legitimate. We take it that this move is unattractive: while it answers the legitimacy question, it does so at the cost of trivialising modest inflationism. 

The challenge, then, is to explain how metaphysical modest inflationism construed along the lines of (b) legitimates ascriptions of mental states and capacities to LLMs while avoiding triviality. The crucial step is to recognise that plausible metaphysically undemanding accounts of mental states---such as a Fodorian analysis of belief---are not arbitrary sets of conditions. Rather, such accounts, even if incomplete as theories of human mental states and capacities, succeed in isolating the criteria for what constitutes a core, objectively specifiable mental kind or mental function.

On this view, an account like Fodor's may be false or incomplete if taken as a total theory of rich, nuanced human belief, which may involve additional layers of self-awareness or emotional texture not captured by the relevant conditions. But this incompleteness for the human case does not preclude the Fodorian criteria from successfully delineating a more fundamental, leaner \textit{type} of representational state that is, in itself, genuinely mental. Let us call such a state `F-belief' (for Fodorian-belief).

If an LLM satisfies the conditions for F-belief, then it is not merely satisfying an arbitrary, stipulated definition. Rather, it is instantiating a mental kind. The ascription of `belief' (qualified, perhaps, as an `F-belief') to the LLM is legitimate precisely because the LLM possesses this objectively specified, genuinely mental (though lean) property. The assertion here, or in relevantly similar cases, is that the LLM instantiates a real, functionally or representationally defined property that falls under the broader category of mentality. 

Thus, modest inflationism posits as a working hypothesis that LLMs can, and sometimes do, instantiate these lean mental kinds. While these kinds may be `metaphysically leaner' than their typical human counterparts (lacking, perhaps, certain experiential or integrative richness), their instantiation by LLMs provides a plausible metaphysical basis for attributing the relevant mental states. This perspective licenses a tempered and cautious form of mentalising---one grounded in evidence for the instantiation of specific, leanly defined mental kinds---rather than unabated anthropomorphism, and it is acutely sensitive to the ways LLM capacities both align with and differ from human cognitive abilities.

\section{Conclusion}

We have argued that inflationism about AI mentality is well-resourced to respond to two deflationist challenges: the Robustness and Etiological Strategies. 

The success of the Robustness Strategy as a method for debunking AI mentality hinges upon substantive assumptions about the nature of mental states and about the kinds of evidence that suffice to demonstrate generalised competence at some task or lack thereof. Accordingly, we have argued that attempts to debunk the existence of mental capacities and capabilities on the grounds of a lack of robustness are weak. With respect to the Etiological Strategy: (1) For etiological arguments from superfluity to succeed, we would need to suppose that mentalistic explanations and mechanistic or functional explanations cannot co-exist and that non-mentalistic explanations are sufficiently informative to have an abductive advantage over mentalistic explanations. We have shown that neither criteria are satisfied by current arguments. (2) In addition, etiological arguments from exclusion require strong claims about what causal history or function is required for the existence of a given mental phenomenon and concrete evidence that this is not present, both of which are contested. Hence the Etiological Strategy fails to provide a strong argument against inflationism.

There may exist other deflationist strategies that we have not considered. It is possible that these strategies can provide a stronger case for deflationism compared to the Robustness and Etiological Strategies. It is also possible that some of the fundamental disputes in cognitive science which underwrite challenges to the Robustness and Etiological Strategies may be resolved in ways that are favourable to the deflationist. Still, it is \textit{at least defensible} to adopt a modest form of inflationism with respect to LLMs. Modest inflationism, so understood, does not default to attributing LLMs the rich mental lives that humans experience, but rather builds its case on the idea that many aspects of mentality---such as beliefs, desires and knowledge states---can be understood in metaphysically undemanding terms, while remaining cautious about more metaphysically demanding aspects of mentality such as phenomenal consciousness. .

\bibliography{references}

\end{document}